\documentclass{ecai_adj}
\usepackage{times}
\usepackage{graphicx}
\usepackage{latexsym}
\usepackage{nicefrac}
\usepackage[algoruled]{algorithm2e}
\usepackage{booktabs}
\usepackage{amsmath}
\usepackage{caption}
\AtBeginDocument{%
   \setlength\abovedisplayskip{3pt}
   \setlength\belowdisplayskip{3pt}}

\begin{document}

\title{Measurable Counterfactual Local Explanations for Any Classifier}

\author{Adam White \and Artur d’Avila Garcez\institute{ City, University of London,
UK, email: Adam.White.2@city.ac.uk A.Garcez@city.ac.uk} }

\maketitle
\bibliographystyle{ecai}

\begin{abstract}
We propose a novel method for explaining the predictions of any classifier. 
In our approach, local explanations are expected to explain both the outcome of a prediction and how that prediction would change if ’things had been different’. Furthermore, we argue that satisfactory explanations cannot be dissociated from a notion and measure of fidelity, as advocated in the early days of neural networks’ knowledge extraction. We introduce a definition of \emph{fidelity to the underlying classifier} for local explanation models which is based on distances to a target decision boundary. A system called CLEAR: Counterfactual Local Explanations via Regression, is introduced and evaluated. CLEAR generates \textit{b}-counterfactual explanations that state minimum changes necessary to flip a prediction's classification. CLEAR then builds local regression models, using the \textit{b}-counterfactuals to measure and improve the fidelity of its regressions. By contrast, the popular LIME method \cite{LIME}, which also uses regression to generate local explanations, neither measures its own fidelity nor generates counterfactuals. CLEAR's regressions are found to have significantly higher fidelity than LIME’s, averaging over 40\% higher in this paper's five case studies.
\end{abstract}

\section{Introduction}
%

Machine learning systems are increasingly being used for automated decision making. It is important that these systems’ decisions can be trusted. This is particularly the case in mission critical situations such as medical diagnosis, airport security or high-value financial trading. Yet the inner workings of many machine learning systems seem unavoidably opaque. The number and complexity of their calculations are often simply beyond the capacities of humans to understand. One possible solution is to treat machine learning systems as ‘black-boxes’ and to then explain their input-output behaviour. Such approaches can be divided into two broad types: those providing global explanations of the entire system and those providing local explanations of single predictions. Local explanations are needed when a machine learning system's decision boundary is too complex to allow for global explanations. This paper focuses on local explanations.

Unfortunately, many explainable AI projects have been too reliant on their researchers' own intuitions as to what constitutes a satisfactory local explanation \cite{miller2018explanation}. Yet the required structure of such explanations has been extensively analysed within philosophy, psychology and cognitive science. Miller\cite{miller2018explanation,miller2018contrastive} has carried out a review of over 250 papers from these disciplines. He states that perhaps his most important finding is that explanations are counterfactual: "they are sought in response to particular counterfactual cases \ldots why event P happened instead of event Q. This has important social and computational consequences for explainable AI". An explanation of a classification needs to show why the machine learning system did not make some alternative (expected or desired) classification.

A novel method called \textbf{C}ounterfactual \textbf{L}ocal \textbf{E}xplanations vi\textbf{A} \textbf{R}egression (CLEAR) is proposed. This is based on the philosopher James Woodard's \cite{Woodward} seminal analysis of counterfactual explanation. Woodward's work derives from Pearl's manipulationlist account of causation \cite{Pearl}. Woodward states that a satisfactory explanation consists in showing patterns of counterfactual dependence. By this he means that it should answer a set of 'what-if-things-had-been-different?' questions, which specify how the explanandum (i.e. the event to be explained) would change if, contrary to the fact, input conditions had been different. It is in this way that a user can understand the relevance of different features, and understand the different ways in which they could change the value of the explanandum. Central to Woodward's notion is the requirement for an explanatory generalization: 
\begin{quote}
"Suppose that M is an explanandum consisting in the statement that some variable Y takes the particular value y. Then an explanans E for M will consist of (a) a generalization G relating changes in the value(s) of a variable X (where X may itself be a vector or n-tuple of variables \(X_i\)) with changes in Y, and (b) a statement (of initial or boundary conditions) that the variable X takes the particular value x."
\end{quote}
In Woodward's analysis, X causes Y. For our purposes, Y can be taken as the machine learning system's predictions where X are the system's input features. The required generalization can be a regression equation that captures the machine learning system's local input-output behaviour. For Woodward, an explanation not only enables an understanding of why an event occurs, it also identifies changes ('manipulations') to features that would have resulted in a different outcome.   

CLEAR provides counterfactual explanations by building on the strengths of two state-of-the-art explanatory methods, while at the same time addressing their weaknesses. The first is by Wachter et al. \cite{Counterfactuals,Explaining} who argue that single predictions are explained by what we shall term as 'boundary counterfactuals' (henceforth: ‘\textit{b}-counterfactuals’) that state the minimum changes needed for an observation to 'flip' its classification. The second method is by Riberio et al. \cite{LIME} who argue for Local Interpretable Model-Agnostic Explanations (LIME). These explanations are created by building a regression model that seeks to approximate the local input-output behaviour of the machine learning system. 

In isolation, \textit{b}-counterfactuals do not provide explanatory generalizations relating X to Y and therefore are not satisfactory explanations, as we exemplify in the next section. LIME, on the other hand, does not measure the fidelity of its regressions and cannot produce counterfactual explanations. 

The contribution of this paper is three-fold. We introduce a novel explanation method capable of: 
\begin{itemize}
  \item providing counterfactuals that are explained by regression coefficients including interaction terms;
  \item evaluating the fidelity of its local explanations to the underlying learning system;
  \item using the values of \textit{b}-counterfactual to significantly improve the fidelity of its regressions.
\end{itemize}

When applied to this paper's five case studies, CLEAR improves on the fidelity of LIME by an average of over 40\%.

Section 2 provides the background to CLEAR including an analysis of \textit{b}-counterfactuals and LIME. Section 3 introduces CLEAR and explains how it uses \textit{b}-counterfactuals to both measure and improve the fidelity of its regressions. Section 4 contains experimental results on five datasets showing that CLEAR's regressions have significantly higher fidelity than LIME's. Section 5 concludes the paper and discusses directions for future work. 

\section{Background}

This paper adopts the following notation: let \textit{m} be a machine learning system mapping \(X\rightarrow Y\); \textit{m} is said to generate prediction \textit{y} for observation \textbf{\textit{x}}. 

\subsection{\textit{b}-Counterfactual Explanations}
\label{headings}
Wachter et al.'s \textit{b}-counterfactuals explain a single prediction by identifying ‘close possible worlds’ in which an individual would receive the prediction they desired.  For example, if a banking machine learning system declined Mr Jones' loan application, a \textit{b}-counterfactual explanation might be that ‘Mr Jones would have received his loan, if his annual salary had been \$35,000 instead of the \$32,000 he currently earns’.  The \$3000 increase would be just sufficient to flip Mr Jones to the desired side of the banking system's decision boundary.

Wachter et al. note that a counterfactual explanation may involve changes to multiple features. Hence, an additional \textit{b}-counterfactual explanation for Mr Jones might be that he would also get the loan if his annual salary was \$33,000 and he had been employed for more than 5 years. Wachter et al. state that counterfactual explanations have the following form:
\begin{quote}
"Score \textit{p} was returned because variables \textit{V} had values ( \(\textit{v}_1,\textit{v}_2, \dots \)) associated with them. If \textit{V} instead had values (\(\textit{v}_1',\textit{v}_2',\dots \)), and all other variables remained constant, score \textit{p}'  would have been returned"  
\end{quote}

\textbf{The key problem with \textit{b}-counterfactuals:} \textit{b}-counterfactual explanations fail to satisfy Woodward's requirement that: \textit{a satisfactory explanation of  prediction y should state a generalization relating X and Y.}

For example, suppose that a machine learning system has assigned Mr Jones a probability of 0.75 for defaulting on a loan. Although stating the changes needed to his salary and years of employment has explanatory value, this falls short of being a satisfactory explanation. A satisfactory explanation also needs to explain:
\begin{enumerate}
\item why Mr Jones was assigned a score of 0.75. This would include identifying the contribution that each feature made to the score.
\item how the features interact with each other. For example, perhaps the number of years employed is only relevant for individuals with salaries below \$34,000.
\end{enumerate}

These requirements could be satisfied by stating an explanatory equation that included interaction terms and indicator variables. At a minimum, the equation's scope should cover a neighbourhood around \textbf{\textit{x}} that includes the data points identified by its \textit{b}-counterfactuals.

\subsection{Local Interpretable Model-Agnostic Explanations}

Ribeiro et al. \cite{LIME} propose LIME, which seeks to explain why \textit{m} predicts \textit{y} given \textbf{\textit{x}} by generating a simple linear regression model that approximates \textit{m}'s input-output behaviour with respect to a small neighbourhood of \textit{m}'s feature space centered on \textbf{\textit{x}}. LIME assumes that for such small neighbourhoods \textit{m}'s input-output behaviour is approximately linear. Ribeiro et al. recognize that there is often a trade off to be made between local fidelity and interpretability. For example, increasing the number of independent variables in a regression might increase local fidelity but decrease interpretability. LIME is becoming an increasingly popular method, and there are now LIME implementations in multiple packages including Python, R and SAS.

\textbf{The LIME algorithm:} Consider a model \textit{m} (e.g. a random forest or MLP) whose prediction is to be explained: The LIME algorithm:
(1) generates a dataset of synthetic observations; (2) labels the synthetic data by passing it to the model \textit{m} which calculates probabilities for each observation belonging to each class.  These probabilities are the ground truths that LIME is trying to explain; 
(3) weights the synthetic observations (in standardised form) using the kernel: \newline 
        \(K(d)=\sqrt{e^{\displaystyle-\left(\nicefrac{d^2}{kernel width^2}\right)}}\)
where \textit{d} is the Euclidean distance from \textbf{\textit{x}} to the synthetic observation, and the default value for kernel width is a function of the number of features in the training dataset; (4) produces a locally weighted regression, using all the synthetic observations. The regression coefficients are meant to explain \textit{m}'s forecast \textit{y}.

\textbf{Key problems with LIME:} \textit{LIME does not measure the fidelity of its regression coefficients. This hides that it may often be producing false explanations.} Although LIME displays the values of its regression coefficients, it does not display the predicted values \textit{y} calculated by its regression model. Let us refer to these values as \emph{regression scores} (they are not bounded by the interval [0,1]).  Ribeiro provides an online 'tutorial' on LIME, which includes an example of a random forest model of the Iris dataset\footnote{https://github.com/marcotcr/lime}. As part of this paper's analysis, the LIME regression scores were captured revealing large errors. For example, in \(\approx\)20\% of explanations, the regression scores differed by more than 0.4 from the probabilities calculated by the random forest model.

\begin{figure*}[t]
   \vspace*{-3mm}
   \caption{Toy example of a machine learning function represented by tan/blue background. The circled cross is \textit{\textbf{x}} whose prediction is to be explained. The other crosses are synthetic observations. (a) LIME uses all synthetic observations in each regression (15,000 in this paper) with weights decreasing with distance from \textit{\textbf{x}}. (b) CLEAR selects a balanced subset of $\approx$ 200 synthetic observation. (c) shows the corresponding b-perturbations. }
  \includegraphics[width=14cm]{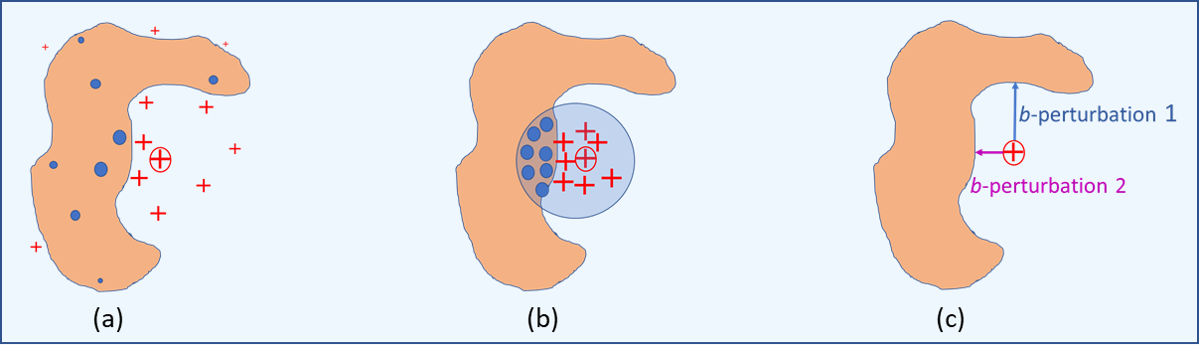}  
  \centering

  \label{boundaries}

\end{figure*}
It might be thought that an adequate solution would be to provide a goodness-of-fit measure such as adjusted R-squared. However, as will be explained in Section 3, such measures can be highly misleading when evaluating the fidelity of the regression coefficients for estimating \textit{b}-counterfactuals.

Another problem is that \textit{LIME does not provide counterfactual explanations.} It might be argued that LIME's regression equations provide the basis for a user to perform their own counterfactual calculations. However, there are multiple reasons why this is incorrect. First, as will be shown in Section 3, additional functionality is necessary for generating faithful counterfactuals including the ability to solve quadratic equations. Second, LIME does not ensure that the regression model correctly classifies \textit{\textbf{x}}. In cases where the regression misclassifies \textbf{\textit{x}}'s class, then any subsequent \textit{b}-counterfactual will be false. Third, it does not have the means of measuring the fidelity of any counterfactual calculations derived from its regression equation. Fourth, LIME does not offer an adequate dataset for calculating counterfactuals. The data that LIME uses in a local regression needs to be representative of the neighbourhood that its explanation is meant to apply to. For counterfactual explanations, this extends from \textit{\textbf{x}} to the nearest points of \textit{m}'s decision boundary. Furthermore, the type of kernel being used is unsuitable; its weightings are too centered around \textbf{\textit{x}}, when other points (e.g. at the decision boundary) are also important.

\subsection{Other Related Work}
Early work seeking to provide explanations to neural networks have been focused on the extraction of symbolic knowledge from trained networks \cite{survey}, either decision trees in the case of feedforward networks \cite{Trepan} or graphs in the case of recurrent networks \cite{Jacobsson,leegiles}. More recently, attention has been shifted from global to local explanation models due to the very large-scale nature of current deep networks, and has been focused on explaining specific network architectures (such as the bottleneck in auto-encoders \cite{Irina}) or domain specific networks such as those used to solve computer vision problems \cite{dissection}, although some recent approaches continue to advocate the use of rule-based knowledge extraction 
\cite{Corels,sontran}. The reader is referred to Guidotti et al. \cite{recentSurvey} for a recent survey.

More specifically, Lundberg et al.\cite{lundberg2017unified} propose SHAP, which explains a prediction by using the game-theory concept of Shapley Values. Shapley Values are the unique solution for fairly attributing the benefits of a cooperative game between players, when subject to a set of local accuracy and consistency constraints. SHAP explains a model m’s prediction for observation \textit{\textbf{x}} by first simplifying \textit{m}’s input data to binary features. A value of 1 corresponds to a feature having the same value as in \textbf{\textit{x}} and a value of 0 corresponds to the feature’s value being ‘unknown’. Within this context, Shapley Values are the changes that each simplified feature makes to \textit{m}’s expected prediction when conditioned on that feature. Lundberg et al. derive a kernel that enables regressions where (i) the dependent variable is \textit{m}’s (rebased) expected prediction conditioned on a specific combination of binary features (ii) the independent features are the binary features (iii) the regression coefficients are the Shapley Values. A key point for this paper is that the Shapley Values apply to the binarized features and cannot be used to estimate the effects of changing a numeric feature of \textbf{\textit{x}} by a particular amount. They therefore do not provide a basis for estimating \textit{b}-counterfactuals. 

Ribeiro et al. \cite{ANCHORS}, the authors of the LIME paper, have subsequently proposed  ‘Anchors: High Precision Model-Agnostic Explanations’.  In motivating their new method they note that LIME does not measure its own fidelity and that 'even the local behaviour of a model may be extremely non-linear, leading to  poor linear approximations'. An Anchor is a rule that is sufficient (with a high probability) to ensure that a local prediction will remain with the same classification, irrespective of the values of other variables. The extent to which the Anchor generalises is estimated by a 'coverage' statistic. For example, an Anchor for a model with the Adult dataset could be: "If Never-Married and Age $\leq$ 28 then \textit{m} will predict '$\leq$ \$50k' with a precision of 97\% and a coverage of 15\%". A pure-exploration multi-armed bandit algorithm is used to efficiently identify Anchors. As with SHAP, Anchors do not provide a basis for estimating \textit{b}-counterfactuals. Therefore neither method can be directly compared to CLEAR.

LIME has spawned several variants. For example LIME-SUP \cite{LIME-SUP} and K-LIME \cite{H20} both seek to explain a machine learning system's functionality over its entire input space by partitioning the input space into a set of neighbourhoods, and then creating local models. K-LIME uses clustering and then regression, LIME-SUP just uses decision tree algorithms. LIME has also been adapted to enable novel applications, for example SLIME \cite{SLIME} provides explanations of sound and music content analysis. However none of these variants address the problems identified with LIME in this paper. 
\begin{figure*}[t]
   \caption{Excerpt from a CLEAR \textit{b}-counterfactual report. In this example CLEAR uses multiple regression to explain a single prediction generated by an MLP model trained on the Pima dataset  }
  \vspace{-3mm} 
  \label{CLEAR output1}
    \centering
  \includegraphics[width=15cm]{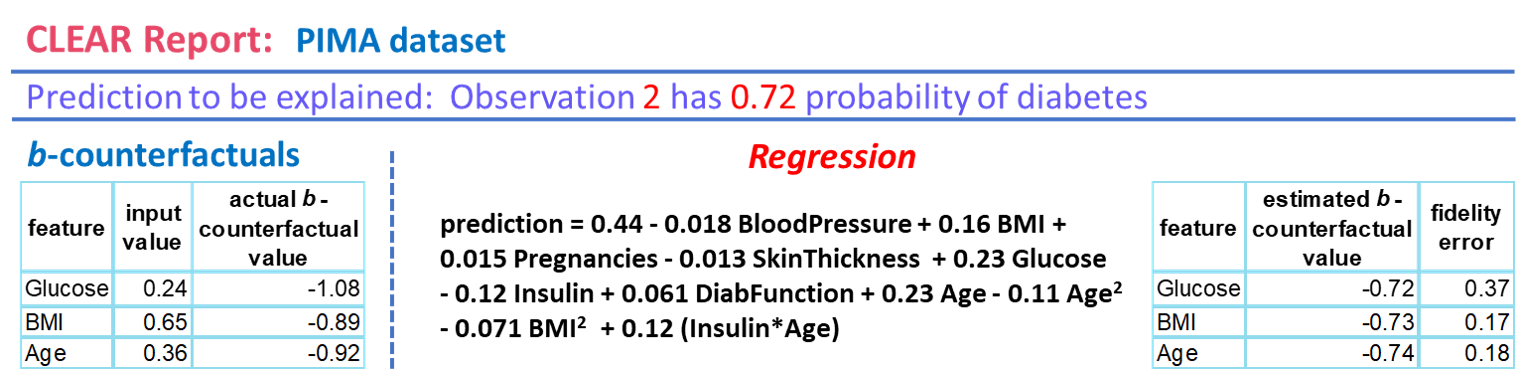}

\end{figure*}
\vspace{5mm}
\section{The CLEAR Method}
CLEAR is based on the view that a satisfactory explanation of a single prediction needs to both explain the value of that prediction and answer 'what-if-things-had-been-different' questions. In doing this it needs to state the relative importance of the input features and show how they interact. A satisfactory explanation must also be measurable and state how well it can explain a model. It must \emph{know when it does not know} \cite{zoubin}.

CLEAR is based on the concept of \textit{b}-perturbation, as follows:

\textit{Definition 5.1} Let min$_f$(\textbf{\textit{x}}) denote a vector resulting from applying a minimum change to the value of one feature $f$ in \textbf{\textit{x}} such that \textit{m}(min$_f$(\textbf{\textit{x}})) = \textit{y'} and \textit{m}(\textbf{\textit{x}}) = \textit{y}, class(\textit{y}) $\neq$ class(\textit{y'}). Let 
$v_f$(\textbf{\textit{x}}) denote the value  of feature $f$ in \textbf{\textit{x}}. A \textbf{\textit{b}-perturbation} is defined as the change in value of feature $f$ for a target class \textit{y'}, that is $v_f$(min$_f$(\textbf{\textit{x}})) $-$ $v_f$(\textbf{\textit{x}}).

For example, for the \textit{b-}counterfactual that Mr Jones would have received his loan if his salary had been \$35,000, a \textit{b}-perturbation for \emph{salary} is \$3000. If \textbf{\textit{x}} has $p$ features and \textit{m} solves a $k$-class problem then there are $q \leq p\times k-1$ \textit{b}-perturbations of \textbf{\textit{x}}; changes in a feature value may not always imply a change of classification.

\begin{figure*}[ht]
\centering
\caption{Comparison of fidelity with different configurations. Fidelity is reduced when any of the above changes are made to the \textit{best configuration}. The changes include:'Imbalanced neighbourhood' = 'using best configuration but with imbalanced neighbourhood datasets', and 'LIME' = 'using the LIME algorithms'. Notice the benefits of CLEAR using a balanced neighbourhood dataset. Also, when CLEAR has 'no quadratic and interaction terms', it is still significantly more faithful than LIME.}
\vspace{-4mm}
\includegraphics[width=15.5cm]{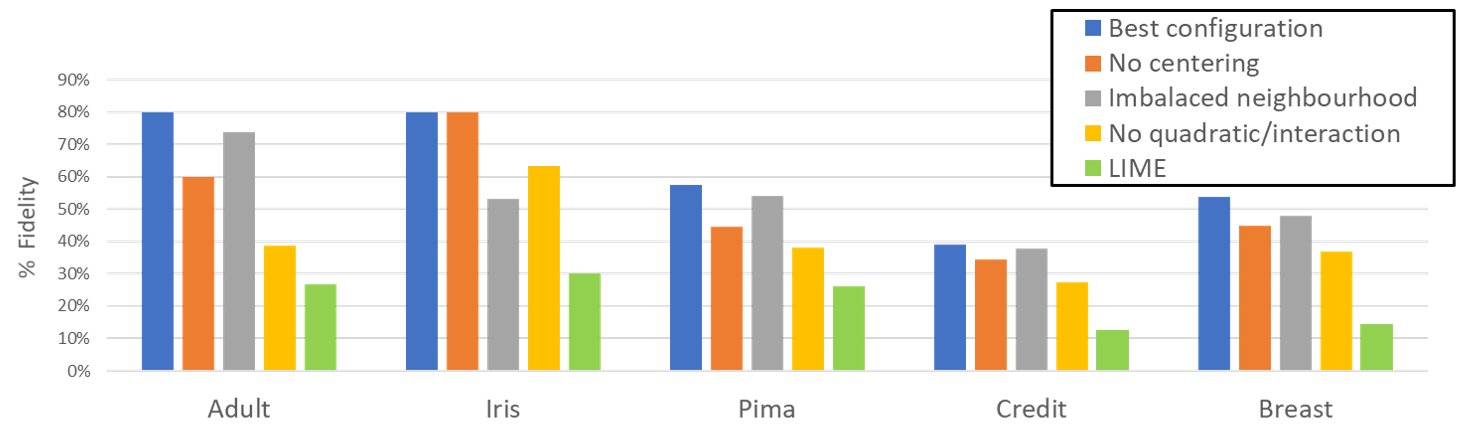}

\label{fidelity configuration}

\end{figure*}

CLEAR compares each \textit{b}-perturbation with an estimate of that value, call it \emph{estimated} \textit{b}-perturbation, calculated using its local regression, to produce a \emph{counterfactual fidelity error}, as follows:
\begin{multline*}
\emph{counterfactual fidelity error} = \\ \mid \text{estimated \textit{b}-perturbation} - \text{\textit{b}-perturbation}\mid
\end{multline*}
Henceforth these will be referred to simply as 'fidelity errors'. CLEAR generates an explanation of prediction \textit{y} for observation \textbf{\textit{x}} by the following steps:

\begin{enumerate}

\item Determine \textbf{\textit{x}}'s \textit{b}-perturbations. For each feature\textit{ f}, a separate one-dimensional search is performed by querying \textit{m} starting at \textit{x}, and progressively moving away from \textit{x} by changing the value of \textit{f} by regular amounts, whilst keeping all other features constant. The searches are constrained to a range of possible feature values.

\item Generate labelled synthetic observations (default: 50,000 observations). Data for numeric features is generated by sampling from a uniform distribution. Data for categorical features is generated by sampling in proportion to the frequencies found in the training set. The synthetic observations are labelled by being passed through \textit{m}.

\item Create a balanced neighbourhood dataset (default is 200 observations). Synthetic observations that are near to \textbf{\textit{x}} (Euclidean distance) are selected with the objective of achieving a dense cloud of points between \textbf{\textit{x}} and the nearest points just beyond \textit{m}'s decision boundaries (Figure \ref{boundaries}). For this, the neighbourhood data is selected such that it is equally distributed across classes, i.e. approximately balanced. CLEAR was also evaluated using  'imbalanced' neighbourhood data, where only the synthetic observations nearest to \textit{\textbf{x}} were selected.  However, it was found this reduced fidelity (see Figure \ref{fidelity configuration}). It might be thought that a balanced neighbourhood dataset should not be used when \textit{\textbf{x}} is far from \textit{m}'s decision boundary; as this would lead to a 'non-local' explanation. But this misses the point that a satisfactory explanation is counterfactual, and that the required locality therefore extends from \textit{\textbf{x}} to data points just beyond \textit{m}'s decision boundary.   

\item Perform a step-wise regression on the neighbourhood dataset, under the constraint that the regression curve should go through \textit{\textbf{x}}. The regression can include second degree terms, interaction terms and indicator variables. CLEAR provides options for both multiple and logistic regression.

\item Estimate the \textit{b}-perturbations by substituting \textbf{\textit{x}}'s \textit{b}-counterfactual values from min$_f$(\textbf{\textit{x}}), other than for feature $f$, into the regression equation and calculating the value of $f$. See example below.

\item Measure the fidelity of the regression coefficients. Fidelity errors are calculated by comparing the actual \textit{b}-perturbations determined in step 1 with the estimates calculated in step 5.

\item Iterate to best explanation. Because CLEAR produces fidelity statistics, its parameters can be iteratively changed in order to achieve a better trade-off between interpretability and fidelity. Relevant parameters include the number of features/independent variables to consider and the use and number of quadratic or interaction terms.
Figure \ref{CLEAR output1} shows excerpts from a CLEAR report.

\item  CLEAR also provides the option of adding \textbf{\textit{x}}'s \textit{b}-counterfactuals, min$_f$(\textbf{\textit{x}}), to \textbf{\textit{x}}'s neighbourhood dataset. The \textit{b}-counterfactuals are weighted and act as soft constraints on CLEAR's subsequent regression. Algorithms 1 and 2 outline the entire process. 
\end{enumerate}

\begin{algorithm}[h]
\SetKwInOut{Input}{input}\SetKwInOut{Output}{output}
\SetAlgoLined
\DontPrintSemicolon
\caption{CLEAR Algorithm}
\Input{ \textit{t }(training data), \textbf{\textit{x}},\textit{m},\textit{T}}
\Output{expl (set of explanations) }
\textit{S}$\leftarrow\ \;$Generate\_Synthetic\_Data(\textbf{\textit{x}},\textit{t},\textit{m})\\ 
\For{each target class tc}{
\For{each feature $f$}{
$w \leftarrow\ \;$Find\_Counterfactuals(\textbf{\textit{x}},\textit{m}) \\
 }
$N_{tc} \leftarrow\;$Balanced\_Neighbourhood(S, \textit{\textbf{x}}, \textit{m})\\ 
Optional: $N_{tc} \leftarrow N_{tc} \cup w$ \\
$r \leftarrow $Find\_Regression\_Equations$(N_{tc},\textit{\textbf{x}})$\\
$w'\leftarrow \;$Estimate\_Counterfactuals(r,\textit{\textbf{x}})\\
$e \leftarrow  \;$Calculate\_Fidelity(w,w',\textit{T})\\
\Return expl$_{tc}=<w,w',r,e>$}
\end{algorithm} 
\begin{algorithm}
\SetKwInOut{Input}{input}\SetKwInOut{Output}{output}
\SetAlgoLined
\DontPrintSemicolon
\caption{Balanced\_Neighbourhood}
\Input{ $S$ (synthetic dataset), \textbf{\textit{x}},\textit{m}\\
$b_1,b_2$ (margins around decision boundary)} 
\Output{$N$ (neighbourhood dataset)}
 $n\leftarrow 200$\\ 
\For{ $s_i \in S$}{
 $d_i \leftarrow$ Euclidean\_Distance $(s_i,\textbf{\textit{x}}$)\;
 $y_i \leftarrow  $m$(s_i)$}
 $N_1 \leftarrow \nicefrac{n}{3}$  members of $ \{S\}\;$ with lowest $d_i$ s.t. $0<y_i \leq b_1$ \;
 $N_2 \leftarrow \nicefrac{n}{3} $  members of $ \{S\}\;$ with lowest $d_i$ s.t. $b_1<y_i \leq b_2$ \;
 $N_3 \leftarrow \nicefrac{n}{3} $  members of $ \{S\}\;$ with lowest $d_i$ s.t. $b_2<y_i \leq 1$ \;
\Return $N \leftarrow N_1 \cup N_2 \cup N_3$ 
\end{algorithm}

Notice that for CLEAR an explanation (expl) is a tuple $<w,w',r,e>$, where $w$ and $w'$ are\textit{ b}-perturbations (actual and estimated), $r$ is a regression equation and $e$ are fidelity errors. 

\begin{table*}[t]
\centering
\caption{Comparison of \% fidelity of CLEAR and LIME: the use of a balanced neighbourhood, centering and quadratic terms allow CLEAR, in general, to achieve a considerably higher fidelity to \textit{b}-counterfactuals than LIME, even without training with \textit{b}-counterfactuals. Including training with \textit{b}-counterfactuals (optional step 8 of CLEAR method), \% fidelity is further increased.}
\begin{tabular}{llllll}
\toprule
 &Adult & Iris & Pima   & Credit & Breast \\
\midrule
CLEAR- not using \textit{b}-counterfactuals & 80\%~{$\pm$}~0.9 & 80\%~{$\pm$}~1.0 & 57\%~{$\pm$}~0.8  &  39\%~{$\pm$}~1.3   &  54\%~{$\pm$}~1.1  \\
CLEAR- using \textit{b}-counterfactuals  & 80\%~{$\pm$}~0.8 & 99.8\%~{$\pm$}~0.1&  77\%~{$\pm$}~0.8  &55\%~{$\pm$}~1.7   &  81\%~{$\pm$}~1.3  \\
LIME algorithms & 26\%~{$\pm$}~0.6 & 30\%~{$\pm$}~0.3 &20\%~{$\pm$}~0.4  & 12\%~{$\pm$}~0.5 & 14\%~{$\pm$}~0.3\\
\bottomrule
\end{tabular}
\end{table*}

   \textbf{Example of using regression to estimate a \textit{b}-perturbation:} An MLP with a softmax activation function in the output layer was trained on a subset of the UCI Pima Indians Diabetes dataset. The MLP calculated a probability of 0.69 for \textbf{\textit{x}} belonging to class 1 (having diabetes). CLEAR generated the logistic regression equation \( (1+e^{\textbf{\textit{w}}^T\textit{\textbf{x}}})^{-1} = 0.69\)\ where:\\
\vspace{-3mm}   
\begin{align*}
\textbf{\textit{\textbf{w}}}^T\textbf{\textit{\textbf{x}}} = -0.8 +1.73\; Glucose + 0.25\;BloodPressure \\ + 0.31\;Glucose^2
\end{align*}

Let the decision boundary be \(P(\textit{\textbf{x}} \in class\ 1) = 0.5\). Thus, \textbf{\textit{x}} is on the boundary when \(\textbf{\textit{w}}^T\textit{\textbf{x }}= 0\). The estimated \textit{b}-perturbation for Glucose is obtained by substituting into the regression equation: \(\textbf{\textit{w}}^T\textbf{\textit{x}} = 0\) and the value of BloodPressure in \textbf{\textit{x}}:
\[-0.31\; Glucose^2 + 1.73\; Glucose -.04 = 0\]
Solving this equation, CLEAR selects the root equal to 0.025 as being closest to the original value of Glucose in \textit{\textbf{x}}. The original value for Glucose was 0.537 and hence the estimated \textit{b}-perturbation is -0.512. The actual \textit{b}-perturbation (step 1) for Glucose to achieve a probability of 0.5 of being in class 1 was -0.557; hence, the fidelity error was 0.045.\\

A CLEAR prototype has been developed in Python\footnote{https://github.com/ClearExplanationsAI}. CLEAR can be run either in batch mode on a test set or it can explain the prediction of a single observation. In batch mode, CLEAR reports the proportion of its estimated \textit{b}-counterfactuals that have a fidelity error lower than a user-specified error threshold \textit{T}, as follows:

\textit{Definition 5.3 (\textbf{\% fidelity}):} A \textit{b}-perturbation is said to be \textit{feasible} if the resulting feature value is within the range of values found in \textit{m}'s training set. The percentage \emph{fidelity} given a batch and error threshold \textit{T} is the number of \textit{b}-perturbations with fidelity error smaller than \textit{T} divided by the number of feasible \textit{b}-perturbations.

Both fidelity and interpretability are critical for a successful explanation. An 'interpretable explanation' that is of poor fidelity is not an explanation, it is just misinformation.  Yet, in order to achieve high levels of fidelity, a CLEAR regression may need to include a large number of terms, including 2nd degree and interaction variables. A criticism of CLEAR might then be that whilst its regression equations are interpretable to data scientists, they may not be interpretable to lay people. However CLEAR's HTML reports are interactive, providing an option to simplify the representation of its equations. Consider Figure \ref{CLEAR output1}, if a user was primarily interested in the effects of Glucose and BMI, then CLEAR can substitute the values of the other features leading to the full regression equation being re-expressed as:
\[ \textbf{\textit{\textbf{w}}}^T\textbf{\textit{\textbf{x}}} = 0.46 +0.3\; Glucose + 0.18\;BMI - 0.05\;BMI^2\]
This equation helps to explain the \textit{b}-counterfactuals, for example it shows why observation 1's classification is more easily 'flipped' by changing the value of Glucose rather than BMI. CLEAR additionally provides options for the user to simplify the original regression, for example by reducing the number of terms, excluding interaction terms etc. CLEAR then enables the user to see the resulting fall in fidelity, putting the user in control of the interpretability/fidelity trade-off. Figure \ref{fidelity configuration} illustrates how fidelity is reduced by excluding both quadratic and interaction terms; notice that even both are excluded, CLEAR's fidelity is much higher than LIME's.
 
\section{Experimental Results}

Experiments were carried out with five UCI datasets: Pima Indians Diabetes (with 8 numeric features), Iris (4 numeric features), Default of Credit Card Clients (20 numeric features, 3 categorical), and subsets of Adult (with 2 numeric features, 5 categorical features),  and Breast Cancer Wisconsin (9 numeric features). For the Adult dataset, some of the categorical features values were merged and features with little predictive power were removed. With the Breast Cancer dataset only the mean value features were kept. For reproducibility, the code for pre-processing the data is included with the CLEAR prototype on GitHub.   

For the Iris dataset, a support vector machine (SVM) with RBF kernel was trained using the scikit-learn library. For each of the other datasets, an MLP with a softmax output layer was trained using Tensorflow. Each dataset was partitioned into a training dataset (used to train the SVM or MLP, and to set the parameters used to generate synthetic data for CLEAR) and a test dataset (out of which 100 observations were selected for calculating the \% fidelity of CLEAR and LIME). Experiments were carried out with different test sets, with each experiment being repeated 20 times for different generated synthetic data. The experiments were carried out on a Windows i7-8700 3.2GHz PC. A single run of a 100 observations took 40-80 minutes, depending on the dataset.

In order to enable comparisons with LIME, CLEAR includes an option to run the LIME algorithms for creating synthetic data and generating regression equations. CLEAR then takes the regression equations and calculates the corresponding \textit{b}-counterfactuals and fidelity errors. It might be objected that it is unfair to evaluate LIME on its counterfactual fidelity, as it was not created for this purpose. But if you accept Woodward and Miller's requirements for a satisfactory explanation, then counterfactual fidelity is an appropriate metric.\\

CLEAR's regressions are significantly better than LIME's. The best results are obtained by including \textit{b}-counterfactuals in the neighbourhood datasets (step 8 of the CLEAR method).
Overall, the \emph{best} configuration comprised: using balanced neighbourhood data, forcing the regression curve to go through \textbf{\textit{x}} (i.e. ’centering’), including both quadratic and interaction terms, and using logistic regression for Pima and Breast Cancer and multiple regression for Iris, Adult and Credit datasets. Unless otherwise stated \% fidelity is for the error threshold \textit{T} = 0.25. 
Table 1 compares the \% fidelity of CLEAR and LIME (i.e. using LIME's algorithms for generating synthetic data and performing the regression). This used LIME's default parameter values except for the following beneficial changes: the number of synthetic data points was increased to 15,000 (further increases did not improve fidelity), the data was not discretized, a maximum of 14 features were allowed, several kernel widths in the range from 1.5 to 4 were evaluated. By contrast, CLEAR was run with its \textit{best} configuration and with 14 features. As an example of LIME's performance: with the Credit dataset, the adjusted \(R^2\) averaged \(\approx\) 0.7, the classification of the test set observations was over 94\% correct. However, the absolute error between \textit{y} and LIME's estimate of \textit{y} was 8\%  (e.g. the MLP forecast \(P(\textit{\textbf{x}} \in class\ 1) = 0.4\), while LIME estimated it at 0.48) and this by itself would lead to large errors when calculating how much a single feature needs to change for \textit{y} to reach the decision boundary. LIME's fidelity of only 12\%, illustrates that CLEAR's measure of fidelity is far more demanding than just classification accuracy. Of course, LIME's poor fidelity was due, in part, to its kernel failing to isolate the appropriate neighbourhood datasets necessary for calculating \textit{b}-counterfactuals accurately.

Table 2 shows how CLEAR's fidelity (not using \textit{b}-counterfactuals) varied with the maximum ’number of independent variables’ allowed in a regression. At first, fidelity sharply improves but then plateaus.

Despite CLEAR's regression fitting the neighbourhood data well, a significant number of the estimated \textit{b}-counterfactuals have large fidelity errors. For example, in one of the experiments with the Adult dataset where the multiple regression did not center the data, the average adjusted \(R^2\) was 0.97, classification accuracy 98\% but the \% fidelity error $<$ 0.25 was 59\%. This points to a more general problem: sometimes the neighbourhood datasets do not represent the regions of its feature space that are central for its explanations. With CLEAR, this discrepancy can at least be measured. 
\begin{table}
\centering

\caption{Variation of \% fidelity with the choice of  number of variables}
  \begin{tabular}{llllll}
   \toprule
   No. & Adult  & Iris   &PIMA     &  Credit & Breast \\
    \midrule
    8 & 35\% & 50\% &42\%  &  27\% & 43\%     \\
    11 & 76\% &62\% &53\%  &  38\% & 46\%           \\
    14 &80\% & 80\% & 57\% & 39\% & 54\% \\
    17& 78\%&  n/a & 59\%  & 40\% & 55\%  \\
    20&78\% &  n/a& 62\%  & 39\% & 56\%  \\
    \bottomrule
 \end{tabular}

\end{table}
%

CLEAR was tested in a variety of configurations. These included the \emph{best} configuration, and configurations where a single option was altered from the default, e.g. by using a imbalanced neighbourhood of points nearest to \textbf{\textit{x}}. Figure \ref{fidelity configuration} displays the results when CLEAR used a maximum of 14 independent variables. 

CLEAR's fidelity was sharply improved by adding \textbf{\textit{x}}'s \textit{b}-counterfactuals to its neighbourhood datasets. In the previous experiments, CLEAR created a neighbourhood dataset of at least 200 synthetic observations, each being given a weighting of 1. This was now altered so that each \textit{b}-counterfactual identified in step 1  was added and given a weighting of 10. For example, for the Pima dataset, an average of \(\approx\)3 \textit{b}-counterfactuals were added to each neighbourhood dataset. The consequent improvement in fidelity indicates as expected that adding these weighted data points results in a dataset capable of representing better the relevant neighbourhood, with CLEAR being able to provide a regression equation that is more faithful to \textit{b}-counterfactuals. 

\section{Conclusion and Future Work}
CLEAR satisfies the requirement that satisfactory local explanations should include statements of key counterfactual cases. CLEAR explains a prediction \textit{y} for data point \textbf{\textit{x}} by stating \textbf{\textit{x}}'s  \textit{b}-counterfactuals and providing a regression equation. The regression shows the patterns of counterfactual dependencies in a neighbourhood that includes both \textbf{\textit{x}} and the \textit{b}-counterfactual data points. CLEAR represents a significant improvement both on LIME and on just using \textit{b}-counterfactuals. Crucial to CLEAR's performance is the ability to generate relevant neighbourhood data bounded by its \textit{b}-counterfactuals. Adding these to \textit{\textbf{x}}'s neighbourhood led to sharp fidelity improvements. Another key feature of CLEAR is that it reports on its own fidelity. Any local explanation system will be fallible, and it is critical with high-value decisions that the user knows if they can trust an explanation. 

There is considerable scope for further developing CLEAR. These include (i) the neighbourhood selection algorithms could be further enhanced. Other data points could also be added, for example \textit{b}-counterfactuals involving changes to multiple features. A user might also include some perturbations that are important to their project. CLEAR could guide this process by reporting regression and fidelity statistics.  And step 8 of the CLEAR algorithm could be replaced by increasingly complex and more sophisticated learning algorithms. Constructing neighbourhood datasets in this way would seem a better approach than randomly generating data points and then selecting those closest to \textbf{\textit{x}} (ii) the search algorithm for actual\textit{ b}-perturbations in step 1 of the CLEAR algorithm could be replaced by a more computationally efficient algorithm (iii). CLEAR should be evaluated in practice in the context of comprehensibility studies.



\begin{quote}
\begin{small}
\bibliography{ecai.bib}
\end{small}
\end{quote}

\end{document}